\pdfoutput=1

\documentclass[11pt]{article}

\usepackage{PRIMEarxiv}
\usepackage[utf8]{inputenc}
\usepackage[T1]{fontenc}
\usepackage{CJKutf8}

\usepackage{hyperref}
\usepackage{microtype}

\usepackage{fancyhdr}
\usepackage{amsmath}
\usepackage{booktabs}
\usepackage{enumitem}
\usepackage{graphicx}
\usepackage{makecell}
\usepackage{multirow}
\usepackage{xurl}

\begin{document}

\title{Overview of the CAIL 2023 Argument Mining Track}

\author{
  Jingcong Liang \\
  Fudan University \\
  \texttt{jcliang22@m.fudan.edu.cn} \\
  \And
  Junlong Wang, Xinyu Zhai \\
  Dalian University of Technology \\
  \texttt{\{jlwang,zhaixy\}@mail.dlut.edu.cn} \\
  \And
  Yungui Zhuang \\
  Tgnet \\
  \texttt{1506025911@qq.com} \\
  \And
  Yiyang Zheng \\
  Shanghai University \\
  \texttt{zhengyiyang@shu.edu.cn} \\
  \And
  Xin Xu \\
  \\
  \texttt{18896502796@163.com} \\
  \And
  Xiandong Ran \\
  Xi'an Jiaotong University \\
  \texttt{rxd1218@stu.xjtu.edu.cn} \\
  \And
  Xiaozheng Dong, Honghui Rong \\
  Lenovo KnowDee Intelligence \\
  \texttt{\{dongxz,ronghh\}@knowdee.com} \\
  \And
  Yanlun Liu \\
  Hunan Uviersity \\
  \texttt{lyl@hnu.edu.cn} \\
  \And
  Hao Chen \\
  Xiamen Huatian International Vocation Institute \\
  \texttt{2390207899@qq.com} \\
  \And
  Yuhan Wei \\
  Northwest Polytechnical University \\
  \texttt{yhwei@mail.nwpu.edu.cn} \\
  \And
  Donghai Li \\
  Beijing Thunisoft Yuandian Information Service Co., Ltd. \\
  \texttt{lidh@thunisoft.com} \\
  \And
  Jiajie Peng \\
  Northwest Polytechnical University \\
  \texttt{jiajiepeng@nwpu.edu.cn} \\
  \And
  Xuanjing Huang \\
  Fudan University \\
  \texttt{xjhuang@fudan.edu.cn} \\
  \And
  Chongde Shi \\
  Beijing Thunisoft Yuandian Information Service Co., Ltd. \\
  \texttt{shichongde@thunisoft.com} \\
  \And
  Yansong Feng \\
  Peking University \\
  \texttt{fengyansong@pku.edu.cn} \\
  \And
  Yun Song\thanks{\enspace Corresponding author.} \\
  Northwest University of Political Science and Law \\
  \texttt{1171991@s.hlju.edu.cn} \\
  \And
  Zhongyu Wei\footnotemark[1] \\
  Fudan University \\
  \texttt{zywei@fudan.edu.cn} \\
}



\maketitle

\begin{abstract}
We give a detailed overview of the CAIL 2023 Argument Mining Track, one of the Chinese AI and Law Challenge (CAIL) 2023 tracks.
The main goal of the track is to identify and extract interacting argument pairs in trial dialogs.
It mainly uses summarized judgment documents but can also refer to trial recordings.
The track consists of two stages, and we introduce the tasks designed for each stage;
we also extend the data from previous events into a new dataset --- CAIL2023-ArgMine --- with annotated new cases from various causes of action.
We outline several submissions that achieve the best results, including their methods for different stages.
While all submissions rely on language models, they have incorporated strategies that may benefit future work in this field.
\keywords{Argument Mining \and Judgment Document Comprehension}
\end{abstract}

\section{Introduction}
\label{sec:introduction}

One of the core tasks for judges during a trial is to understand the arguments and testimonies from the two sides, as they need to base their verdicts on them appropriately.
Typically, the panel summarizes the trial dialog into a judgment document, on which this judicial argument comprehension process heavily depends~\cite{Vermeule1999Judicial};
however, the panel has to perform the task manually by reading the entire document.
The call for automating this process and other judicial tasks has yielded various studies and models involving both statistical methods~\cite{Ulmer1963Quantitative,Nagel1963Applying} and natural language processing (NLP) techniques~\cite{Sulea2017Exploring,Katz2017general,Liu2018two}, including the latest large language models~\cite{Cui2023ChatLaw,Yue2023DISC,Mou2024Unveiling,Zhang2024SoMeLVLM}.

However, judicial argument comprehension differs from other tasks in the judicial field, as it requires extracting arguments from the trial dialog (and its summarized form) and understanding their interaction, especially between sides.
The arguing procedure greatly resembles general argumentation procedures like debates, where such automatic comprehension is also possible~\cite{Lin2023Argue,Liang2024Debatrix}.
More specifically, the judicial argument comprehension task largely overlaps the domain of argument mining~\cite{Lawrence2019Argument}.
The development of NLP has also resulted in many studies that aimed to automate argument mining, covering argument structure prediction~\cite{Wang2011Predicting,Stab2014Identifying,Liu2019Discourse} and interaction identification~\cite{Taghipour2016Neural,Wei2016Is,Tan2016Winning,Habernal2016Which,Dong2017Attention,Ji2018Incorporating,Cheng2020APE,Ji2021Discrete}.

Based on the tasks designed by~\cite{Ji2021Discrete}, \cite{Yuan2021Overview} hosted the SMP-CAIL2020-Argmine Challenge\footnote{\url{http://cail.cipsc.org.cn/task_summit.html?raceID=3&cail_tag=2020}}, a track of the Chinese AI and Law Challenge (CAIL) 2020.
The track addressed the importance of adapting argument-mining techniques to judicial scenarios and assisting judicial argument comprehension.
Participants must develop models to identify and extract \emph{interacting arguments} in trial dialogs.
Here, two arguments interact if they come from different sides (plaintiff and defense) and make claims about the same subject, as shown in Table~\ref{tab:int_arg_example}.
Interacting arguments can be consistent or partially consistent, yet in most cases, one disputes another.

\begin{CJK*}{UTF8}{gbsn}

\begin{table*}
    \centering
    \caption{Examples of interacting argument pairs between the plaintiff and defense in different trials. Note that in the last example, the arguments are partially consistent.}\label{tab:int_arg_example}
    \begin{tabular}{ccp{0.8\linewidth}}
\toprule
No.                & Side      & Argument                                                                                          \\ \midrule
\multirow{2}{*}{1} & Plaintiff & 同年3月14日凌晨，被告人何**回到家中向高*谎称李某某走失，并阻止高*报警但未果。                     \\
                   & Defense   & 当晚11点至12点其回家后，便与女儿高*报警。                                                         \\ \midrule
\multirow{2}{*}{2} & Plaintiff & 被告在家很少操持家务，也很少做饭。                                                                \\
                   & Defense   & 被告一直在家伺候公婆和孩子的生活起居，还要兼顾浴池的生意。                                        \\ \midrule
\multirow{2}{*}{3} & Plaintiff & 2015年12月26日，被告王**、周**因果园种植需要，从原告处借款3万元，约定月利率1.5％，借款期限6个月。 \\
                   & Defense   & 被告王**借原告款属实，但借款时原告当场扣除了3000元。                                              \\ \bottomrule
\end{tabular}
\end{table*}

\end{CJK*}

As the direct successor to the SMP-CAIL2020-Argmine Challenge, the latest CAIL 2023 Argument Mining Track\footnote{\url{http://cail.cipsc.org.cn/task_summit.html?raceID=5&cail_tag=2023}} extended the datasets and introduced new tasks closer to real scenarios.
We divided the track into two stages, examining interacting argument identification and extraction abilities.
More specifically, the first stage requires the model to choose the argument from five candidates by the defense that interacts with the given plaintiff's argument.
The second stage requires the model to extract interacting argument pairs from judgment documents directly.

In this paper, we report and analyze the best submissions to the track, including their methods and evaluation measures in both stages.
All these submissions used language models as their backbone, yet they have used different approaches to improve their performance further.
We also discuss the findings and reflections from these submissions for future study in this area.

\section{Related Work}
\label{sec:related}

Automatic processing of judicial tasks, such as judgment document analysis, has been researched since the 1950s.
Early studies mainly focused on quantitative and statistical methods~\cite{Kort1957Predicting,Ulmer1963Quantitative,Nagel1963Applying};
more recent works also introduced machine learning approaches~\cite{Lauderdale2012Supreme}.
Meanwhile, the rapid development of natural language processing (NLP) has greatly aided the current progress in this field.
For example,~\cite{Sulea2017Exploring} and~\cite{Liu2018two} trained \(N\)-gram based classifiers for judicial verdict prediction;
\cite{Katz2017general} and~\cite{Luo2017Learning} utilized various features to predict criminal case charges.
The latest studies also utilized the power of large language models (LLM) to tackle general judicial tasks and queries~\cite{Cui2023ChatLaw,Yue2023DISC}.
In 2018,~\cite{Zhong2018Overview} held the Chinese AI Law Challenge (CAIL), initially focusing on judgment prediction, along with a large-scale legal dataset~\cite{Xiao2018CAIL2018}.
Since then, more judicial tasks and datasets were included, such as similar case matching~\cite{Xiao2019CAIL2019} and judicial document classification~\cite{Liu2006Exploring}.

The specific task of argumentative analysis of judgment documents was introduced in 2020 for its unique characteristic primarily related to argument mining~\cite{Yuan2021Overview}.
Argument mining research focuses on defining and evaluating new algorithms for identifying and structurally analyzing human argumentation.
Typical tasks in this area include argument segmentation and extraction~\cite{Levy2014Context,Levy2017Unsupervised,Ajjour2017Unit,Mou2022Two}, argument classification~\cite{Stab2014Identifying,Niculae2017Argument,ToledoRonen2020Multilingual} and argument relation mining~\cite{Peldszus2015Joint,Persing2016End,Bao2021Neural}.
However, the most related task to the CAIL 2023 Argument Mining Track is interactive argument detection, a combination of argument extraction and relation mining~\cite{ElBaff2020Analyzing,Cheng2020APE}.
Specifically, \cite{Ji2021Discrete} proposed identifying interacting argument pairs in online debate forums, which was the direct inspiration for the tasks in our track.
Since then, various methods have been proposed to tackle this task \cite{Yuan2021Leveraging,Liang2023Hi}.

\section{Task Details}
\label{sec:task}

\subsection{Task Description}

The CAIL 2023 Argument Mining Track focused on recognizing interacting argument pairs in trial dialogs.
As mentioned in Section~\ref{sec:introduction}, we define interacting argument pairs as two arguments, by the plaintiff and defense, respectively, that argue over the same subject.
A typical judgment document lists the plaintiff's claims and evidence first, followed by the defense's rebuttals.
Therefore, in most cases, each of the defense's arguments is recorded as acknowledging or denying one or more arguments by the plaintiff, forming one or more interacting argument pairs.
As a result, the track aimed to find these interacting argument pairs in judgment documents.

We split the track into two stages.
The first stage requires identifying pairing arguments from the given candidates, and the second requires extracting argument pairs from judgment documents.
Due to the low density of interacting argument pairs, argument pair extraction is more complicated than identification.
In this way, we can distinguish models with various levels of ability.

More specifically, in the first stage, given an argument by the plaintiff, we provide five candidate arguments sampled from the defense side.
One of these arguments forms an interacting argument paired with the plaintiff's argument.
We guarantee that only one argument can form such a pair, and the model should identify and choose it.
Since this task only involves pre-extracted arguments from the judgment document, it is relatively easy but not very practical in real scenarios.

In the second stage, we only provide the judgment documents;
the model must extract all interacting argument pairs between the plaintiff and defense.
This task is more difficult since only a few arguments can form such pairs, causing the labels to be highly imbalanced.
We also provided trial recordings as multimodal data for a portion of the documents as a bonus;
these recordings were transcribed, with the speaker of each speech annotated.
Samples with and without trial recordings are separated, resulting in two subtasks evaluated separately (using the same metrics).

\subsection{Dataset}

In the SMP-CAIL2020-Argmine Challenge,~\cite{Yuan2021Overview} annotated a dataset with \(4,476\) interacting argument pairs from \(1,069\) judgment documents.
Their annotation included the case type (civil, criminal, etc.), the cause of action, the plaintiff and defense, and a list of interacting argument pairs.
They also filtered cases that lacked substantial argumentation or were too short or long.
In CAIL 2021, they extended the original dataset with more cases.
However, the original dataset and the extended version lacked variety regarding the cause of action, especially civil ones.

Following their annotation schema, we refined their dataset and extended it with more trial cases from various crime types. Details of the data collection and annotation procedure are available in Appendix~\ref{sec:annotation}, including annotation statistics. We name the new dataset CAIL2023-Argmine, and we list its data statistics in Table~\ref{tab:data_stat_overall}.
CAIL2023-Argmine is twice as large as the original dataset and covers more causes of action, especially civil cases.
The argument pair density remains low, again emphasizing the challenge of the track tasks.

\begin{table}
    \centering
    \caption{Overall statistics of the original dataset and CAIL2023-Argmine. 2021: Extended version of the original dataset for CAIL 2021; 2021*: The previous dataset without cases not belonging to any specific cause of action (the last three statistics also exclude them); 2023: CAIL2023-Argmine.}\label{tab:data_stat_overall}
    \begin{tabular}{l@{}rrr}
\toprule
\multicolumn{1}{c}{Statistics} & \multicolumn{1}{c}{2021} & \multicolumn{1}{c}{2021*} & \multicolumn{1}{c}{2023} \\ \midrule
\# Documents                   & \(1,872\)                & \(1,449\)                 & \(3,620\)                \\
\# Argument Pairs              & \(10,077\)               & \(7,618\)                 & \(20,009\)               \\
Pair Density (\%)              & \(6.60\)                 & \(6.20\)                  & \(8.88\)                 \\
\# Causes of Actions           & \(5\)                    & \(5\)                     & \(13\)                   \\
\# Criminal Cases              & \(1,304\)                & \(1,304\)                 & \(2,291\)                \\
\# Civil Cases                 & \(145\)                  & \(145\)                   & \(1,329\)                \\ \bottomrule
\end{tabular}
\end{table}

\begin{CJK*}{UTF8}{gbsn}

Table~\ref{tab:cause_stat} lists data statistics per cause of action in CAIL2023-Argmine.
While the cases in the original dataset mainly came from a few causes of action like the crime of intentional injury (``故意伤害罪'') or causing traffic casualties (``交通肇事罪''), new cases are distributed more uniformly and cover more causes of action.
Furthermore, the original dataset only contains one civil cause of action --- maritime disputes (``海事海商纠纷'') --- with only \(145\) documents.
In CAIL2023-Argmine, we annotated nearly \(1,200\) new documents from \(4\) different civil causes of action, greatly extending this part of the data.
Another observation is that the argument pair density in the new cases is higher than in the old ones, especially in criminal cases.

\begin{table}
    \centering
    \caption{Data statistics per cause of action. 2021: the extended version of the dataset in the SMP-CAIL2020-Argmine Challenge; 2023*: new cases included in CAIL2023-Argmine. Causes of action that end with ``罪'' are criminal causes, while others are civil ones. Note that ``其他'' is not a valid cause of action, and we excluded corresponding cases in CAIL2023-Argmine.}\label{tab:cause_stat}
    \begin{tabular}{ll@{}rrrr}
\toprule
\multicolumn{1}{c}{Dataset} & \multicolumn{1}{c}{Cause of Action} & \multicolumn{1}{c}{\# Docs} & \multicolumn{1}{c}{\# Sentences} & \multicolumn{1}{c}{\# Pairs} & \multicolumn{1}{c}{Density (\%)} \\ \midrule
\multirow{6}{*}{2021}       & 故意伤害罪                               & \(956\)                          & \(16,601\)                       & \(4,183\)                    & \(5.44\)                         \\
                            & 交通肇事罪                               & \(315\)                          & \(5,237\)                        & \(2,394\)                    & \(10.52\)                        \\
                            & 故意杀人罪                               & \(29\)                           & \(405\)                          & \(152\)                      & \(10.22\)                        \\
                            & 虐待罪                                 & \(4\)                            & \(101\)                          & \(30\)                       & \(6.13\)                         \\ \cmidrule(l){2-6} 
                            & 海事海商纠纷                              & \(145\)                          & \(2,639\)                        & \(859\)                      & \(7.59\)                         \\ \cmidrule(l){2-6} 
                            & 其他*                                 & \(423\)                          & \(7,922\)                        & \(2,459\)                    & \(6.20\)                         \\ \midrule
\multirow{9}{*}{2023*}      & 故意杀人罪                               & \(272\)                          & \(4,005\)                        & \(1,366\)                    & \(9.25\)                         \\
                            & 抢劫罪                                 & \(253\)                          & \(3,579\)                        & \(1,698\)                    & \(13.41\)                        \\
                            & 强奸罪                                 & \(239\)                          & \(3,271\)                        & \(1,501\)                    & \(13.22\)                        \\
                            & 诈骗罪                                 & \(168\)                          & \(2,601\)                        & \(1,255\)                    & \(12.54\)                        \\
                            & 盗窃罪                                 & \(55\)                           & \(800\)                          & \(469\)                      & \(16.23\)                        \\ \cmidrule(l){2-6} 
                            & 借款合同纠纷                              & \(299\)                          & \(4,157\)                        & \(1,353\)                    & \(9.27\)                         \\
                            & 离婚纠纷                                & \(298\)                          & \(4,612\)                        & \(1,498\)                    & \(8.23\)                         \\
                            & 生命权、身体权、健康权纠纷                       & \(294\)                          & \(4,515\)                        & \(1,727\)                    & \(9.73\)                         \\
                            & 买卖合同纠纷                              & \(293\)                          & \(4,248\)                        & \(1,524\)                    & \(9.88\)                         \\ \bottomrule
\end{tabular}
\end{table}

\end{CJK*}

In addition to judgment documents, we prepared trial audio and transcriptions for the multimodal bonus task in the second stage.
Due to limited open access to trial recordings, we could only obtain trial videos for some cases in CAIL2023-Argmine.
We extracted the audio from the videos and transcribed them into dialog texts using speech recognition models.
Since the model output did not contain speaker information, we further annotated the speaker of each speech in the dialog.
Table~\ref{tab:multi_data_stat} shows statistics of these multimodal data.

\begin{table}[t]
    \centering
    \caption{Statistics of multimodal data in CAIL2023-Argmine. There are more recordings than documents because some cases have multiple trial recordings.}\label{tab:multi_data_stat}
    \begin{tabular}{lr}
\toprule
\multicolumn{1}{c}{Statistics} & \multicolumn{1}{c}{Value} \\ \midrule
\# Documents                   & \(134\)                   \\
\# Argument Pairs              & \(688\)                   \\
Pair Density (\%)              & \(8.49\)                  \\ \midrule
\# Recordings                  & \(146\)                   \\
Average Duration               & \(50'12''866\)            \\
\# Speeches                    & \(22,654\)                \\ \bottomrule
\end{tabular}
\end{table}

\subsection{Evaluation and Baseline}

For the first stage, we evaluate the submissions by calculating the accuracy of candidate prediction over the test set:
\begin{equation}
    S_1=\mathrm{Acc}=\frac{\sum_{i=1}^nI\{y_i=\hat{y}_i\}}{n}
\end{equation}
where \(y_i\) and \(\hat{y}_i\) are the true label and prediction of the \(i\)-th sample in the test set respectively.

However, the accuracy was no longer suitable for the second stage due to imbalanced labels (pairing or not), as seen in Table~\ref{tab:data_stat_overall}.
Therefore, we used the \(F_1\) score as the evaluation metric for this stage:
\begin{equation}
    F_1=\frac{2\mathrm{TP}}{2\mathrm{TP}+\mathrm{FP}+\mathrm{FN}}
\end{equation}
where \(\mathrm{TP}\) is the number of correct predictions for \emph{all} cases, \(\mathrm{FP}\) is the number of incorrect predictions, and \(\mathrm{FN}\) is the number of missed pairs.
Specifically, we calculated \(F_1\) scores for multimodal and text-only samples separately, obtaining \(F_1^\mathrm{m}\) and \(F_1^\mathrm{t}\) respectively.
We took their weighed average as the score of this stage:
\begin{equation}
    S_2=0.8F_1^\mathrm{t}+0.2F_1^\mathrm{m}
\end{equation}

Finally, we ranked teams that participated in both stages using the following integrated score:
\begin{equation}
    S=0.3S_1+0.7S_2
\end{equation}

We provided a BERT-based baseline model~\cite{Devlin2019BERT} for both stages during the track\footnote{\url{https://github.com/china-ai-law-challenge/CAIL2023/tree/main/lblj}}.
This baseline model treats both tasks as binary classification.
It pairs the plaintiff's argument to each of the defense's candidate arguments and predicts whether they form an interacting argument pair or not;
Specifically, it does not use judgment documents, trial transcripts, or multimodal data.
We trained the model with the training set of each stage to obtain the final baseline for that stage.

In the first stage, the model achieved an accuracy of \(0.60\);
however, in the second stage, it only achieved \(F_1^\mathrm{m}=0.42\) and \(F_1^\mathrm{t}=0.30\).
The evaluation result further proved that the task in the second stage is more difficult than in the first stage.

\section{Track Results}
\label{sec:track}

In the CAIL 2023 Argument Mining Track, \(35\) teams from universities and enterprises have submitted their models during the first stage.
Among them, \(28\) teams successfully superseded our baseline model and were able to participate in the second stage.
Finally, \(10\) teams have submitted valid models for the second stage.

\subsection{Submissions}
\label{subsec:submissions}

This section provides descriptions of the top \(7\) teams to whom we awarded prizes.
The rest of the teams were not required to submit technical reports about their models, yet we will still show their results in Section~\ref{subsec:results}.

\textbf{DUT-large} fine-tuned multiple pre-trained language models and employed a soft voting approach to ensemble them.
In the first stage, they concatenated the plaintiff's argument with each of the defense's candidate arguments as the model input.
To enhance robustness, they introduced adversarial training through Projected Gradient Descent (PGD) and R-Drop regularization.
In the second stage, they used a similar approach by treating the task as a binary classification task.
They involved sentence-BERT models~\cite{Reimers2019Sentence} for this stage.
They employed several techniques, such as pseudo-labeling, focal loss, and the Fast Gradient Method (FGM) for adversarial training.

\textbf{X} used BERT-based models for both stages but under different approaches.
In the first stage, they fine-tuned a MacBERT model~\cite{Cui2021Pre} integrated with a multi-choice setup to match arguments.
They also leveraged the sentence-BERT model for sentence pair binary classification in the second stage.

\textbf{zyy} utilized Qwen-14B~\cite{Bai2023Qwen} for both stages.
The model was fine-tuned by Low-Rank Adaptation (LoRA)~\cite{Hu2022LoRA}, with NefTune applied to enhance performance.
They created prompts for each stage, respectively.
In the first stage, the prompt included the plaintiff's argument and all five candidates and instructed the model to choose the interacting one.
In the second stage, they provided arguments from both sides, prompting the model to find all matching pairs between them.

\textbf{xxxin} fine-tuned a MacBERT model for the first stage.
In addition to argument pairs, they included designated tokens to identify the cause of action (including its category and specific cause) and the plaintiff's and defense's arguments.
Each batch consisted of argument pairs with the same plaintiff's argument but with different defense arguments.
In the second stage, they used multiple methods, such as pseudo-labels, negative sampling, and adversarial training, to deal with imbalanced labels.

\textbf{AG} adopted a list-wise learning approach, transforming both stages into a sorting task.
In the first stage, they used the provided candidates directly as a list, and in the second stage, they constructed the list using Cartesian products.
They fine-tuned the Lawformer model~\cite{Xiao2021Lawformer}, utilizing adversarial learning with FGM.

\begin{CJK*}{UTF8}{gkai}

\textbf{福气by} introduced multi-model voting for both stages.
In the first stage, they employed a combination of a multiple choice question-answering model, a text-matching model capturing argument relation, and a re-ranking model to refine the selection.
They did not include the question-answering model and the re-ranking model for the second stage;
instead, they enhanced the text-matching model with negative samples.

\textbf{酸菜饺子\&玉米饺子} fine-tuned a RoBERTa~\cite{Liu2019RoBERTa} model on large law corpora.
They also adopted FGM to enhance the robustness of the model.

\end{CJK*}

\subsection{Results}
\label{subsec:results}

\begin{CJK*}{UTF8}{gbsn}

Table~\ref{tab:result} demonstrates the CAIL 2023 Argument Mining Track results, including the scores of both stages and the final score to rank teams.
Although all teams performed better than the baseline in the first stage, not all could beat the baseline in the second stage.
DUT-large significantly outperformed all other teams in both stages, followed by {X} and {zyy}, which are very close.
Other teams mentioned in Section~\ref{subsec:submissions} were on par with X and {zyy} in one stage but not in the other (``福气by'' and ``酸菜饺子\&玉米饺子'' in the first, {xxxin} and AG in the second).

\begin{table}
    \centering
    \caption{CAIL 2023 Argument Mining Track final results, sorted by the final score \(S\). Only teams with valid submissions for the second stage are listed. We round all scores to two digits.}\label{tab:result}
    \begin{tabular}{lrrrc}
\toprule
\multicolumn{1}{c}{Team} & \multicolumn{1}{c}{\(S_1\)} & \multicolumn{1}{c}{\(S_2\)} & \multicolumn{1}{c}{\(S\)} & Rank   \\ \midrule
DUT-large                & \(0.72\)                    & \(0.50\)                    & \(0.56\)                  & \(1\)  \\
X                        & \(0.68\)                    & \(0.45\)                    & \(0.52\)                  & \(2\)  \\
zyy                      & \(0.69\)                    & \(0.44\)                    & \(0.52\)                  & \(3\)  \\
xxxin                    & \(0.63\)                    & \(0.45\)                    & \(0.51\)                  & \(4\)  \\
AG                       & \(0.65\)                    & \(0.44\)                    & \(0.50\)                  & \(5\)  \\
福气by                   & \(0.69\)                    & \(0.42\)                    & \(0.50\)                  & \(6\)  \\
酸菜饺子\&玉米饺子       & \(0.68\)                    & \(0.35\)                    & \(0.45\)                  & \(7\)  \\
婷之队                   & \(0.67\)                    & \(0.30\)                    & \(0.41\)                  & \(8\)  \\
争议观点队               & \(0.63\)                    & \(0.28\)                    & \(0.39\)                  & \(9\)  \\
sxu-wzt                  & \(0.64\)                    & \(0.09\)                    & \(0.26\)                  & \(10\) \\ \bottomrule
\end{tabular}
\end{table}

\end{CJK*}

\begin{figure}
    \centering
    \includegraphics[width=0.5\linewidth]{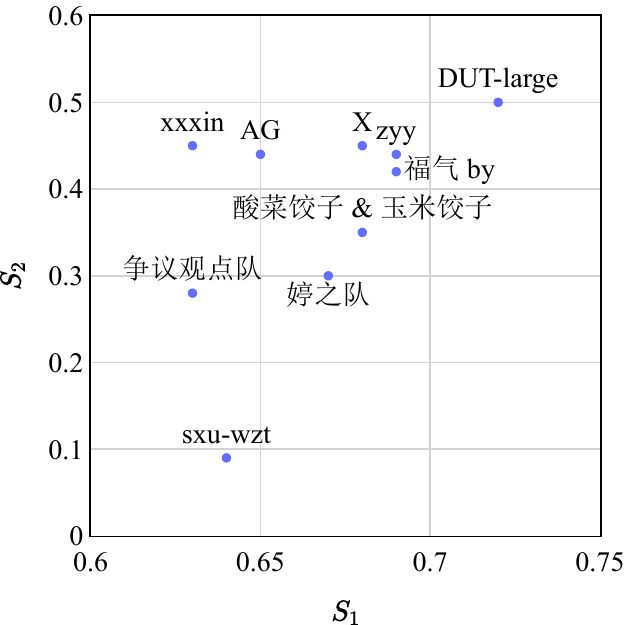}
    \caption{Score distribution of teams with valid submissions for the second stage.}
    \label{fig:scatter}
\end{figure}

The correlation coefficient of \(S_1\) and \(S_2\) is \(0.5191\), and the distribution of the two scores of different teams is illustrated in Figure~\ref{fig:scatter}.
Although most teams shared models across the two stages, the scores of the two tasks are not strictly correlated.
This is probably because the two tasks, though deeply related, vary in some aspects such as the density of interactive argument pairs.

\section{Analysis}
\label{sec:analysis}

\subsection{Common Techniques}

Upon analyzing the models proposed by top teams, we discovered some common techniques that many teams have adopted:

\paragraph{Task Transformation}
In both stages, all teams have transformed the task into more standard ones.
In the first stage, teams treated the task as binary classification or multiple choice question answering.
In the second stage, binary classification was more prevalent, while some teams also turned the task into a ranking problem.
Such transformation allowed teams to adopt mature solutions for predefined standard tasks that are easier to implement and train.

\paragraph{Pretrained Language Model}
All teams have utilized pre-trained language models (PLM) as their backbone, reflecting the power of such models in NLP tasks.
The selected models varied among teams, from smaller BERT-sized models (sBERT, RoBERTa) to larger LLMs (Qwen-14B).
Some teams introduced models specifically pre-trained on law corpora  (Lawformer) or more general Chinese corpora (ERNIE)~\cite{Sun2021ERNIE} to match the tasks more closely.

\paragraph{Domain Specific Fine-tuning}
Regardless of the backbone PLM selection, all teams have fine-tuned their backbones on the training set we provided and other similar data.
This domain adaptation approach could enhance the proficiency of PLMs within the specified scenario and task.
Teams adopted different input/output formats and training strategies depending on the model used and how they transformed the tasks.

\subsection{Promising Tricks}

Besides the above methods, teams have proposed different tricks to improve their results.
We list some of them below:

\paragraph{Data Augmentation}
Some teams have involved data augmentation techniques, including pseudo-labeling and negative sampling.
These methods could alleviate the negative effect of imbalanced labels, especially in the second stage.
Meanwhile, they could improve the generalizability and robustness of the model.

\begin{CJK*}{UTF8}{gbsn}

\paragraph{Adversarial Training}
Besides data augmentation, several teams have used adversarial training to fight against imbalanced labels.
For instance, DUT-large, AG, and ``酸菜饺子\&玉米饺子'' used the Fast Gradient Method (FGM) to implement adversarial training;
Moreover, DUT-large adopted Projected Gradient Descent (PGD) for adversarial training in the first stage.

\end{CJK*}

\paragraph{Loss Function and Regularization}
Since most teams treated the tasks as binary classification, they generally chose cross entropy as the loss function.
However, other loss functions were also introduced, such as focal loss.
Teams also applied regularization methods like R-Drop to enhance model robustness.

\paragraph{Inter-task Transfer Learning}
Several teams exploited their model from the first stage during the second stage, transferring it to the new scenario.
For example, AG unified the tasks from the two stages and performed further fine-tuning to adapt to the second stage.

\paragraph{Large Langauge Model}
As mentioned, {zyy} utilized Qwen-14B, a large language model (LLM) for text generation.
Therefore, they have involved unique fine-tuning strategies for LLMs such as LoRA and NefTune.
They also designed prompt templates for the task in the two stages.

\paragraph{Model Ensembling}
Many teams trained multiple models and compared their performance during each stage.
Several teams have employed model ensembling to exploit all model outputs for further improvements.
They designed different methods for the final prediction, such as direct or soft voting.

\begin{CJK*}{UTF8}{gbsn}

\begin{table}
    \centering
    \caption{\(F_1\) scores of the two subtasks in the second stage. We only include teams whose model performed better (achieving a higher \(S_2\)) than our baseline in this stage.}\label{tab:stage_2_result}
    \begin{tabular}{lrrr}
\toprule
\multicolumn{1}{c}{Team} & \multicolumn{1}{c}{\(F_1^\mathrm{t}\)} & \multicolumn{1}{c}{\(F_1^\mathrm{m}\)} & \multicolumn{1}{c}{\(S_2\)} \\ \midrule
DUT-large                & \(0.52\)                               & \(0.44\)                               & \(0.50\)                    \\
xxxin                    & \(0.47\)                               & \(0.39\)                               & \(0.45\)                    \\
X                        & \(0.47\)                               & \(0.37\)                               & \(0.45\)                    \\
zyy                      & \(0.45\)                               & \(0.42\)                               & \(0.44\)                    \\
AG                       & \(0.46\)                               & \(0.36\)                               & \(0.44\)                    \\
福气by                   & \(0.44\)                               & \(0.33\)                               & \(0.42\)                    \\ \midrule
Baseline                 & \(0.42\)                               & \(0.30\)                               & \(0.39\)                    \\ \bottomrule
\end{tabular}
\end{table}

\end{CJK*}

\subsection{Multimodal Bonus Task}

As stated in Section~\ref{sec:task}, we split out a multimodal bonus task in the second stage, providing trial recordings and transcriptions for the documents in this task.
However, none of the top \(7\) teams listed in Section~\ref{subsec:submissions} utilized the multimodal data, only extracting interacting argument pairs from the given judgment document.
We conjecture that the large gap between judgment documents and trial recordings (including transcriptions) could have made it more difficult to exploit the multimodal resource.

Nevertheless, Table~\ref{tab:stage_2_result} shows the \(F_1\) scores of teams listed in Table~\ref{tab:result} that superseded our baseline in the second stage.
Most teams utilized BERT-like PLMs and followed a similar score distribution as our baseline, where \(F_1^\mathrm{m}\) is significantly lower than \(F_1^\mathrm{t}\).
This phenomenon follows our expectation because the cases from the bonus task had different causes of action compared to other cases in this stage.
In contrast, zyy has gained an \(F_1^\mathrm{m}\) much closer to \(F_1^\mathrm{t}\), using a true LLM (Qwen-14B).
This phenomenon suggests that modern LLMs are more robust in cross-domain tasks.

\section{Conclusion}
\label{sec:conclusion}

This paper presented the CAIL 2023 Argument Mining Track, a track of the Chinese AI and Law Challenge (CAIL) 2023 that focused on identifying and extracting interacting argument pairs in judgment documents.
We divided the track into two consecutive stages;
we received \(10\) valid submissions for the second stage from \(28\) teams that passed the threshold in the first stage.
We introduced the tasks and corresponding data for each stage and outlined the best submissions.
Compared to our provided baseline, teams have generally performed better, using various techniques that are beneficial for similar tasks in judicial scenarios.

However, there is still a long way to achieve the goal of fully automated judicial argument comprehension.
The tasks we proposed in the two stages are simplified from real-world demands, potentially limiting their application in actual trials.
Moreover, it remains unclear how much multimodal data from trial recordings can assist such tasks, as the top teams did not exploit the multimodal data we provided.

Nevertheless, we thank all participating teams for taking the time to participate in this challenging track, especially the top teams in the second stage that provided technical reports of their submissions.
We believe that future research can use the tasks, data, and submissions from our track to build more powerful automatic systems for judicial use.

\section*{Acknowledgements}
This work is supported by National Natural Science Foundation of China (No. 62176058) and National Key R \& D Program of China (2023YFF1204800). The project's computational resources are supported by CFFF platform of Fudan University.

\bibliographystyle{unsrt}
\bibliography{main}

\begin{thebibliography}{10}

\bibitem{Vermeule1999Judicial}
Adrian Vermeule.
\newblock Judicial history.
\newblock {\em The Yale Law Journal}, 108(6):1311--1354, 1999.

\bibitem{Ulmer1963Quantitative}
S.~Sidney Ulmer.
\newblock Quantitative analysis of judicial processes: Some practical and theoretical applications.
\newblock {\em Law and Contemporary Problems}, 28(1):164--184, 1963.

\bibitem{Nagel1963Applying}
Stuart~S. Nagel.
\newblock Applying correlation analysis to case prediction.
\newblock {\em Texas Law Review}, 42:1006, 1963.

\bibitem{Sulea2017Exploring}
Octavia-Maria Sulea, Marcos Zampieri, Shervin Malmasi, Mihaela Vela, Liviu~P. Dinu, and Josef van Genabith.
\newblock Exploring the use of text classification in the legal domain.
\newblock October 2017.

\bibitem{Katz2017general}
Daniel~Martin Katz, Michael~J. Bommarito, II, and Josh Blackman.
\newblock A general approach for predicting the behavior of the supreme court of the united states.
\newblock {\em PLOS ONE}, 12(4):1--18, 04 2017.

\bibitem{Liu2018two}
Yi-Hung Liu and Yen-Liang Chen.
\newblock A two-phase sentiment analysis approach for judgement prediction.
\newblock {\em Journal of Information Science}, 44(5):594--607, 2018.

\bibitem{Cui2023ChatLaw}
Jiaxi Cui, Zongjian Li, Yang Yan, Bohua Chen, and Li~Yuan.
\newblock Chatlaw: Open-source legal large language model with integrated external knowledge bases.
\newblock June 2023.

\bibitem{Yue2023DISC}
Shengbin Yue, Wei Chen, Siyuan Wang, Bingxuan Li, Chenchen Shen, Shujun Liu, Yuxuan Zhou, Yao Xiao, Song Yun, Xuanjing Huang, and Zhongyu Wei.
\newblock Disc-lawllm: Fine-tuning large language models for intelligent legal services.
\newblock September 2023.

\bibitem{Mou2024Unveiling}
Xinyi Mou, Zhongyu Wei, and Xuanjing Huang.
\newblock Unveiling the truth and facilitating change: Towards agent-based large-scale social movement simulation.
\newblock February 2024.

\bibitem{Zhang2024SoMeLVLM}
Xinnong Zhang, Haoyu Kuang, Xinyi Mou, Hanjia Lyu, Kun Wu, Siming Chen, Jiebo Luo, Xuanjing Huang, and Zhongyu Wei.
\newblock Somelvlm: A large vision language model for social media processing.
\newblock February 2024.

\bibitem{Lin2023Argue}
Jiayu Lin, Rong Ye, Meng Han, Qi~Zhang, Ruofei Lai, Xinyu Zhang, Zhao Cao, Xuanjing Huang, and Zhongyu Wei.
\newblock Argue with me tersely: Towards sentence-level counter-argument generation.
\newblock In Houda Bouamor, Juan Pino, and Kalika Bali, editors, {\em Proceedings of the 2023 Conference on Empirical Methods in Natural Language Processing}, pages 16705--16720, Singapore, December 2023. Association for Computational Linguistics.

\bibitem{Liang2024Debatrix}
Jingcong Liang, Rong Ye, Meng Han, Ruofei Lai, Xinyu Zhang, Xuanjing Huang, and Zhongyu Wei.
\newblock Debatrix: Multi-dimensional debate judge with iterative chronological analysis based on llm.
\newblock March 2024.

\bibitem{Lawrence2019Argument}
John Lawrence and Chris Reed.
\newblock Argument mining: A survey.
\newblock {\em Computational Linguistics}, 45(4):765--818, December 2019.

\bibitem{Wang2011Predicting}
Li~Wang, Marco Lui, Su~Nam Kim, Joakim Nivre, and Timothy Baldwin.
\newblock Predicting thread discourse structure over technical web forums.
\newblock In Regina Barzilay and Mark Johnson, editors, {\em Proceedings of the 2011 Conference on Empirical Methods in Natural Language Processing}, pages 13--25, Edinburgh, Scotland, UK., July 2011. Association for Computational Linguistics.

\bibitem{Stab2014Identifying}
Christian Stab and Iryna Gurevych.
\newblock Identifying argumentative discourse structures in persuasive essays.
\newblock In {\em Proceedings of the 2014 Conference on Empirical Methods in Natural Language Processing ({EMNLP})}, pages 46--56, Doha, Qatar, October 2014. Association for Computational Linguistics.

\bibitem{Liu2019Discourse}
Jiangming Liu, Shay~B. Cohen, and Mirella Lapata.
\newblock Discourse representation parsing for sentences and documents.
\newblock In Anna Korhonen, David Traum, and Llu{\'\i}s M{\`a}rquez, editors, {\em Proceedings of the 57th Annual Meeting of the Association for Computational Linguistics}, pages 6248--6262, Florence, Italy, July 2019. Association for Computational Linguistics.

\bibitem{Taghipour2016Neural}
Kaveh Taghipour and Hwee~Tou Ng.
\newblock A neural approach to automated essay scoring.
\newblock In Jian Su, Kevin Duh, and Xavier Carreras, editors, {\em Proceedings of the 2016 Conference on Empirical Methods in Natural Language Processing}, pages 1882--1891, Austin, Texas, November 2016. Association for Computational Linguistics.

\bibitem{Wei2016Is}
Zhongyu Wei, Yang Liu, and Yi~Li.
\newblock Is this post persuasive? ranking argumentative comments in online forum.
\newblock In Katrin Erk and Noah~A. Smith, editors, {\em Proceedings of the 54th Annual Meeting of the Association for Computational Linguistics (Volume 2: Short Papers)}, pages 195--200, Berlin, Germany, August 2016. Association for Computational Linguistics.

\bibitem{Tan2016Winning}
Chenhao Tan, Vlad Niculae, Cristian Danescu-Niculescu-Mizil, and Lillian Lee.
\newblock Winning arguments: Interaction dynamics and persuasion strategies in good-faith online discussions.
\newblock In {\em Proceedings of the 25th International Conference on World Wide Web}, WWW '16, page 613–624, Republic and Canton of Geneva, CHE, 2016. International World Wide Web Conferences Steering Committee.

\bibitem{Habernal2016Which}
Ivan Habernal and Iryna Gurevych.
\newblock Which argument is more convincing? analyzing and predicting convincingness of web arguments using bidirectional {LSTM}.
\newblock In Katrin Erk and Noah~A. Smith, editors, {\em Proceedings of the 54th Annual Meeting of the Association for Computational Linguistics (Volume 1: Long Papers)}, pages 1589--1599, Berlin, Germany, August 2016. Association for Computational Linguistics.

\bibitem{Dong2017Attention}
Fei Dong, Yue Zhang, and Jie Yang.
\newblock Attention-based recurrent convolutional neural network for automatic essay scoring.
\newblock In Roger Levy and Lucia Specia, editors, {\em Proceedings of the 21st Conference on Computational Natural Language Learning ({C}o{NLL} 2017)}, pages 153--162, Vancouver, Canada, August 2017. Association for Computational Linguistics.

\bibitem{Ji2018Incorporating}
Lu~Ji, Zhongyu Wei, Xiangkun Hu, Yang Liu, Qi~Zhang, and Xuanjing Huang.
\newblock Incorporating argument-level interactions for persuasion comments evaluation using co-attention model.
\newblock In Emily~M. Bender, Leon Derczynski, and Pierre Isabelle, editors, {\em Proceedings of the 27th International Conference on Computational Linguistics}, pages 3703--3714, Santa Fe, New Mexico, USA, August 2018. Association for Computational Linguistics.

\bibitem{Cheng2020APE}
Liying Cheng, Lidong Bing, Qian Yu, Wei Lu, and Luo Si.
\newblock {APE}: Argument pair extraction from peer review and rebuttal via multi-task learning.
\newblock In {\em Proceedings of the 2020 Conference on Empirical Methods in Natural Language Processing (EMNLP)}, pages 7000--7011, Online, November 2020. Association for Computational Linguistics.

\bibitem{Ji2021Discrete}
Lu~Ji, Zhongyu Wei, Jing Li, Qi~Zhang, and Xuanjing Huang.
\newblock Discrete argument representation learning for interactive argument pair identification.
\newblock In {\em Proceedings of the 2021 Conference of the North American Chapter of the Association for Computational Linguistics: Human Language Technologies}, pages 5467--5478, Online, June 2021. Association for Computational Linguistics.

\bibitem{Yuan2021Overview}
Jian Yuan, Zhongyu Wei, Yixu Gao, Wei Chen, Yun Song, Donghua Zhao, Jinglei Ma, Zhen Hu, Shaokun Zou, Donghai Li, and Xuanjing Huang.
\newblock {Overview of SMP-CAIL2020-Argmine: The Interactive Argument-Pair Extraction in Judgement Document Challenge}.
\newblock {\em Data Intelligence}, 3(2):287--307, June 2021.

\bibitem{Kort1957Predicting}
Fred Kort.
\newblock Predicting supreme court decisions mathematically: A quantitative analysis of the “right to counsel” cases.
\newblock {\em American Political Science Review}, 51(1):1–12, 1957.

\bibitem{Lauderdale2012Supreme}
B.~E. Lauderdale and T.~S. Clark.
\newblock The supreme court’s many median justices.
\newblock {\em American Political Science Review}, 106(4):847–866, 2012.

\bibitem{Luo2017Learning}
Bingfeng Luo, Yansong Feng, Jianbo Xu, Xiang Zhang, and Dongyan Zhao.
\newblock Learning to predict charges for criminal cases with legal basis.
\newblock In Martha Palmer, Rebecca Hwa, and Sebastian Riedel, editors, {\em Proceedings of the 2017 Conference on Empirical Methods in Natural Language Processing}, pages 2727--2736, Copenhagen, Denmark, September 2017. Association for Computational Linguistics.

\bibitem{Zhong2018Overview}
Haoxi Zhong, Chaojun Xiao, Zhipeng Guo, Cunchao Tu, Zhiyuan Liu, Maosong Sun, Yansong Feng, Xianpei Han, Zhen Hu, Heng Wang, and Jianfeng Xu.
\newblock Overview of cail2018: Legal judgment prediction competition.
\newblock October 2018.

\bibitem{Xiao2018CAIL2018}
Chaojun Xiao, Haoxi Zhong, Zhipeng Guo, Cunchao Tu, Zhiyuan Liu, Maosong Sun, Yansong Feng, Xianpei Han, Zhen Hu, Heng Wang, and Jianfeng Xu.
\newblock Cail2018: A large-scale legal dataset for judgment prediction.
\newblock July 2018.

\bibitem{Xiao2019CAIL2019}
Chaojun Xiao, Haoxi Zhong, Zhipeng Guo, Cunchao Tu, Zhiyuan Liu, Maosong Sun, Tianyang Zhang, Xianpei Han, Zhen Hu, Heng Wang, and Jianfeng Xu.
\newblock Cail2019-scm: A dataset of similar case matching in legal domain.
\newblock November 2019.

\bibitem{Liu2006Exploring}
Chao-Lin Liu and Chwen-Dar Hsieh.
\newblock Exploring phrase-based classification of judicial documents for criminal charges in chinese.
\newblock In {\em Proceedings of the 16th International Conference on Foundations of Intelligent Systems}, ISMIS'06, page 681–690, Berlin, Heidelberg, 2006. Springer-Verlag.

\bibitem{Levy2014Context}
Ran Levy, Yonatan Bilu, Daniel Hershcovich, Ehud Aharoni, and Noam Slonim.
\newblock Context dependent claim detection.
\newblock In Junichi Tsujii and Jan Hajic, editors, {\em Proceedings of {COLING} 2014, the 25th International Conference on Computational Linguistics: Technical Papers}, pages 1489--1500, Dublin, Ireland, August 2014. Dublin City University and Association for Computational Linguistics.

\bibitem{Levy2017Unsupervised}
Ran Levy, Shai Gretz, Benjamin Sznajder, Shay Hummel, Ranit Aharonov, and Noam Slonim.
\newblock Unsupervised corpus{--}wide claim detection.
\newblock In Ivan Habernal, Iryna Gurevych, Kevin Ashley, Claire Cardie, Nancy Green, Diane Litman, Georgios Petasis, Chris Reed, Noam Slonim, and Vern Walker, editors, {\em Proceedings of the 4th Workshop on Argument Mining}, pages 79--84, Copenhagen, Denmark, September 2017. Association for Computational Linguistics.

\bibitem{Ajjour2017Unit}
Yamen Ajjour, Wei-Fan Chen, Johannes Kiesel, Henning Wachsmuth, and Benno Stein.
\newblock Unit segmentation of argumentative texts.
\newblock In Ivan Habernal, Iryna Gurevych, Kevin Ashley, Claire Cardie, Nancy Green, Diane Litman, Georgios Petasis, Chris Reed, Noam Slonim, and Vern Walker, editors, {\em Proceedings of the 4th Workshop on Argument Mining}, pages 118--128, Copenhagen, Denmark, September 2017. Association for Computational Linguistics.

\bibitem{Mou2022Two}
Xinyi Mou, Zhongyu Wei, Changjian Jiang, and Jiajie Peng.
\newblock A two stage adaptation framework for frame detection via prompt learning.
\newblock In Nicoletta Calzolari, Chu-Ren Huang, Hansaem Kim, James Pustejovsky, Leo Wanner, Key-Sun Choi, Pum-Mo Ryu, Hsin-Hsi Chen, Lucia Donatelli, Heng Ji, Sadao Kurohashi, Patrizia Paggio, Nianwen Xue, Seokhwan Kim, Younggyun Hahm, Zhong He, Tony~Kyungil Lee, Enrico Santus, Francis Bond, and Seung-Hoon Na, editors, {\em Proceedings of the 29th International Conference on Computational Linguistics}, pages 2968--2978, Gyeongju, Republic of Korea, October 2022. International Committee on Computational Linguistics.

\bibitem{Niculae2017Argument}
Vlad Niculae, Joonsuk Park, and Claire Cardie.
\newblock Argument mining with structured {SVM}s and {RNN}s.
\newblock In {\em Proceedings of the 55th Annual Meeting of the Association for Computational Linguistics (Volume 1: Long Papers)}, pages 985--995, Vancouver, Canada, July 2017. Association for Computational Linguistics.

\bibitem{ToledoRonen2020Multilingual}
Orith Toledo-Ronen, Matan Orbach, Yonatan Bilu, Artem Spector, and Noam Slonim.
\newblock Multilingual argument mining: Datasets and analysis.
\newblock In {\em Findings of the Association for Computational Linguistics: EMNLP 2020}, pages 303--317, Online, November 2020. Association for Computational Linguistics.

\bibitem{Peldszus2015Joint}
Andreas Peldszus and Manfred Stede.
\newblock Joint prediction in {MST}-style discourse parsing for argumentation mining.
\newblock In Llu{\'\i}s M{\`a}rquez, Chris Callison-Burch, and Jian Su, editors, {\em Proceedings of the 2015 Conference on Empirical Methods in Natural Language Processing}, pages 938--948, Lisbon, Portugal, September 2015. Association for Computational Linguistics.

\bibitem{Persing2016End}
Isaac Persing and Vincent Ng.
\newblock End-to-end argumentation mining in student essays.
\newblock In {\em Proceedings of the 2016 Conference of the North {A}merican Chapter of the Association for Computational Linguistics: Human Language Technologies}, pages 1384--1394, San Diego, California, June 2016. Association for Computational Linguistics.

\bibitem{Bao2021Neural}
Jianzhu Bao, Chuang Fan, Jipeng Wu, Yixue Dang, Jiachen Du, and Ruifeng Xu.
\newblock A neural transition-based model for argumentation mining.
\newblock In {\em Proceedings of the 59th Annual Meeting of the Association for Computational Linguistics and the 11th International Joint Conference on Natural Language Processing (Volume 1: Long Papers)}, pages 6354--6364, Online, August 2021. Association for Computational Linguistics.

\bibitem{ElBaff2020Analyzing}
Roxanne El~Baff, Henning Wachsmuth, Khalid Al~Khatib, and Benno Stein.
\newblock {A}nalyzing the {P}ersuasive {E}ffect of {S}tyle in {N}ews {E}ditorial {A}rgumentation.
\newblock In {\em Proceedings of the 58th Annual Meeting of the Association for Computational Linguistics}, pages 3154--3160, Online, July 2020. Association for Computational Linguistics.

\bibitem{Yuan2021Leveraging}
Jian Yuan, Zhongyu Wei, Donghua Zhao, Qi~Zhang, and Changjian Jiang.
\newblock Leveraging argumentation knowledge graph for interactive argument pair identification.
\newblock In Chengqing Zong, Fei Xia, Wenjie Li, and Roberto Navigli, editors, {\em Findings of the Association for Computational Linguistics: ACL-IJCNLP 2021}, pages 2310--2319, Online, August 2021. Association for Computational Linguistics.

\bibitem{Liang2023Hi}
Jingcong Liang, Rong Ye, Meng Han, Qi~Zhang, Ruofei Lai, Xinyu Zhang, Zhao Cao, Xuanjing Huang, and Zhongyu Wei.
\newblock Hi-{A}r{G}: Exploring the integration of hierarchical argumentation graphs in language pretraining.
\newblock In Houda Bouamor, Juan Pino, and Kalika Bali, editors, {\em Proceedings of the 2023 Conference on Empirical Methods in Natural Language Processing}, pages 14606--14620, Singapore, December 2023. Association for Computational Linguistics.

\bibitem{Devlin2019BERT}
Jacob Devlin, Ming-Wei Chang, Kenton Lee, and Kristina Toutanova.
\newblock {BERT}: Pre-training of deep bidirectional transformers for language understanding.
\newblock In Jill Burstein, Christy Doran, and Thamar Solorio, editors, {\em Proceedings of the 2019 Conference of the North {A}merican Chapter of the Association for Computational Linguistics: Human Language Technologies, Volume 1 (Long and Short Papers)}, pages 4171--4186, Minneapolis, Minnesota, June 2019. Association for Computational Linguistics.

\bibitem{Reimers2019Sentence}
Nils Reimers and Iryna Gurevych.
\newblock Sentence-{BERT}: Sentence embeddings using {S}iamese {BERT}-networks.
\newblock In Kentaro Inui, Jing Jiang, Vincent Ng, and Xiaojun Wan, editors, {\em Proceedings of the 2019 Conference on Empirical Methods in Natural Language Processing and the 9th International Joint Conference on Natural Language Processing (EMNLP-IJCNLP)}, pages 3982--3992, Hong Kong, China, November 2019. Association for Computational Linguistics.

\bibitem{Cui2021Pre}
Yiming Cui, Wanxiang Che, Ting Liu, Bing Qin, and Ziqing Yang.
\newblock Pre-training with whole word masking for chinese bert.
\newblock {\em IEEE/ACM Transactions on Audio, Speech, and Language Processing}, 29:3504--3514, 2021.

\bibitem{Bai2023Qwen}
Jinze Bai, Shuai Bai, Yunfei Chu, Zeyu Cui, Kai Dang, Xiaodong Deng, Yang Fan, Wenbin Ge, Yu~Han, Fei Huang, Binyuan Hui, Luo Ji, Mei Li, Junyang Lin, Runji Lin, Dayiheng Liu, Gao Liu, Chengqiang Lu, Keming Lu, Jianxin Ma, Rui Men, Xingzhang Ren, Xuancheng Ren, Chuanqi Tan, Sinan Tan, Jianhong Tu, Peng Wang, Shijie Wang, Wei Wang, Shengguang Wu, Benfeng Xu, Jin Xu, An~Yang, Hao Yang, Jian Yang, Shusheng Yang, Yang Yao, Bowen Yu, Hongyi Yuan, Zheng Yuan, Jianwei Zhang, Xingxuan Zhang, Yichang Zhang, Zhenru Zhang, Chang Zhou, Jingren Zhou, Xiaohuan Zhou, and Tianhang Zhu.
\newblock Qwen technical report.
\newblock September 2023.

\bibitem{Hu2022LoRA}
Edward~J. Hu, Yelong Shen, Phillip Wallis, Zeyuan Allen-Zhu, Yuanzhi Li, Shean Wang, Lu~Wang, and Weizhu Chen.
\newblock Lo{RA}: Low-rank adaptation of large language models.
\newblock In {\em International Conference on Learning Representations}, 2022.

\bibitem{Xiao2021Lawformer}
Chaojun Xiao, Xueyu Hu, Zhiyuan Liu, Cunchao Tu, and Maosong Sun.
\newblock Lawformer: A pre-trained language model for chinese legal long documents.
\newblock {\em AI Open}, 2:79--84, 2021.

\bibitem{Liu2019RoBERTa}
Yinhan Liu, Myle Ott, Naman Goyal, Jingfei Du, Mandar Joshi, Danqi Chen, Omer Levy, Mike Lewis, Luke Zettlemoyer, and Veselin Stoyanov.
\newblock Roberta: A robustly optimized bert pretraining approach.
\newblock July 2019.

\bibitem{Sun2021ERNIE}
Yu~Sun, Shuohuan Wang, Shikun Feng, Siyu Ding, Chao Pang, Junyuan Shang, Jiaxiang Liu, Xuyi Chen, Yanbin Zhao, Yuxiang Lu, Weixin Liu, Zhihua Wu, Weibao Gong, Jianzhong Liang, Zhizhou Shang, Peng Sun, Wei Liu, Xuan Ouyang, Dianhai Yu, Hao Tian, Hua Wu, and Haifeng Wang.
\newblock Ernie 3.0: Large-scale knowledge enhanced pre-training for language understanding and generation.
\newblock July 2021.

\end{thebibliography}

\appendix

\section{Annotation Details}
\label{sec:annotation}

\begin{CJK*}{UTF8}{gbsn}

The dataset for SMP-CAIL2020-Argmine (including its extension in 2021) only covered \(5\) causes of action.
Meanwhile, many cases did not have a specific cause of action (tagged as ``其他'').
First, we excluded the above vague cases in CAIL2023-Argmine to ensure the quality of crime features.
Next, to increase the diversity of the dataset, we selected \(9\) causes of action that had no or few cases in the original dataset, including criminal and civil ones, as listed in Table~\ref{tab:anno_stat}.
We collected public judgment documents and filtered them according to the rules proposed by~\cite{Yuan2021Overview};
for each cause of action, if there are more than \(300\) valid cases, we sampled \(300\) of them.

\begin{table}
    \centering
    \caption{Statistics of the annotation for new cases in CAIL2023-Argmine. Causes of action that end with ``罪'' are criminal causes, while others are civil ones. ``Agreement'' and ``Disagreement'' denote agreement and disagreement argument pairs, respectively. We define agreement and disagreement pairs as in~\cite{Yuan2021Overview}: we only count argument pairs where the defense fully acknowledges the plaintiff as agreement ones.}\label{tab:anno_stat}
    \begin{tabular}{l@{}rrrr}
\toprule
\multicolumn{1}{c}{Cause of Action} & \multicolumn{1}{c}{\# Documents} & \multicolumn{1}{c}{\# Agreement} & \multicolumn{1}{c}{\# Disagreement} & \multicolumn{1}{c}{Cohen's \(\kappa\)} \\ \midrule
故意杀人罪                          & \(272\)                          & \(604\)                            & \(762\)                               & \(0.18\)                               \\
抢劫罪                              & \(253\)                          & \(794\)                            & \(904\)                               & \(0.26\)                               \\
强奸罪                              & \(239\)                          & \(414\)                            & \(1,087\)                             & \(0.28\)                               \\
诈骗罪                              & \(168\)                          & \(602\)                            & \(653\)                               & \(0.25\)                               \\
盗窃罪                              & \(55\)                           & \(319\)                            & \(150\)                               & \(0.21\)                               \\ \midrule
借款合同纠纷                        & \(299\)                          & \(246\)                            & \(1,107\)                             & \(0.15\)                               \\
离婚纠纷                            & \(298\)                          & \(282\)                            & \(1,216\)                             & \(0.24\)                               \\
生命权、身体权、健康权纠纷          & \(294\)                          & \(82\)                             & \(1,645\)                             & \(0.28\)                               \\
买卖合同纠纷                        & \(293\)                          & \(209\)                            & \(1,315\)                             & \(0.22\)                               \\ \midrule
Summary                             & \(2,171\)                        & \(3,552\)                          & \(8,839\)                             & \(0.24\)                               \\ \bottomrule
\end{tabular}
\end{table}

\end{CJK*}

Next, we arranged for a group of undergraduate and graduate students majoring in law to annotate these documents.
We only kept the part containing testimonies from both sides and extracted the sentences from the document beforehand to unify the argument span.
The annotation schema was the same as in~\cite{Yuan2021Overview}, where annotators needed to choose the case type, the cause of action, the plaintiff and defense, and a list of interacting argument pairs, except that they only needed to select pre-extracted sentences as argument candidates.
Two annotators annotated each document.
If the list of selected argument pairs differed between the annotators, an adjudicator would decide on the valid pairs among all candidates.
We invited a law professor as the adjudicator to ensure the annotation quality.

\begin{CJK*}{UTF8}{gbsn}

After annotating all documents, we further filtered documents whose annotated cause of action (by both annotators) did not match its presumed one because it indicates that the cause of action could be vague in this case.
Table~\ref{tab:anno_stat} shows annotation statistics of the final documents per cause of action.
In most cases, there are more disagree pairs than agree pairs, especially in civil cases;
the only exception is lancery (``盗窃罪''), where the evidence is usually undoubtedly clear.
The Cohen's \(\kappa\) is not very high, making adjudication critical during annotation.

\end{CJK*}

\end{document}